\documentclass{article}

 \usepackage[preprint]{neurips_2026}

\usepackage[utf8]{inputenc} 
\usepackage[T1]{fontenc}    
\usepackage{hyperref}       
\usepackage{url}            
\usepackage{booktabs}       
\usepackage{amsfonts}       
\usepackage{nicefrac}       
\usepackage{microtype}      
\usepackage{xcolor}         
\usepackage{graphicx}
\title{SCAFFOLD: A Large-Scale Structured Dataset of Computer Science Research Figures with Diagram QA and Chain-of-Thought Reasoning Traces}

\author{%
  Ranjit Raut \\
  Department of Artificial Intelligence\\
  Kathmandu University\\
  Dhulikhel, Nepal \\
  \And
  Aarav Subedi \\
  Department of Artificial Intelligence\\
  Kathmandu University\\
  Dhulikhel, Nepal \\
  \And
  Sagun Rai \\
  Department of Artificial Intelligence\\
  Kathmandu University\\
  Dhulikhel, Nepal \\
  \And
  Sudan Jha \\
  Department of Computer Science and Engineering\\
  Kathmandu University\\
  Dhulikhel, Nepal \\
}

\begin{document}

\maketitle

\begin{abstract}
Computer science papers rely heavily on diagrams: architecture drawings, system flowcharts, and pipeline schematics, that often carry more information than the text around them. There is currently no public dataset that pairs this specific kind of figure with captions, context, questions, answers, and step-by-step reasoning, which is exactly what is needed to train a vision-language model to understand them. We present \textbf{SCAFFOLD}\footnote{https://github.com/theranjitraut/scaffold}, a large-scale structured dataset of computer science research figures with diagram QA and Chain-of-Thought reasoning traces. This dataset consists of (image, caption, context, question-answer, chain-of-thought) tuples from arXiv computer science papers prepared using layout detection and PDF parsing, with an AI-assisted question-generation step. The resulting large-sized SCAFFOLD-157K dataset spans 3,058 papers with 29,887 figures (157,387 pairs), a medium-sized SCAFFOLD-37K dataset (36,797 pairs), and a small-sized SCAFFOLD-12K dataset (12,000 pairs). We used SCAFFOLD-12K for baseline experiment on Qwen2.5-VL-3B-Instruct.
\end{abstract}

\section{Introduction}

Computer science papers are full of visual content. A single architecture diagram, flowchart, or pipeline schematic can encode relationships that would otherwise take several paragraphs of text to explain. Yet the tools researchers use to search and process the literature -- search engines, citation graphs, text summarizers -- mostly ignore this visual content. If a reader wants to understand what a diagram is actually saying, they still have to work it out by eye.

Recent vision-language models (VLMs) can process images and text together, and general-purpose systems like LLaVA, BLIP-2, and PaliGemma do well on everyday visual question answering. But these models are rarely tested -- and rarely trained -- on the specific visual language of computer science papers: boxes-and-arrows architecture diagrams, multi-stage pipelines, and flowcharts where the right answer depends on tracing a particular connection or step order, not just recognizing objects in a photo.

This gap exists not because models lack capability, but because the data doesn't exist. No public dataset pairs computer science diagrams with captions, context, questions, and reasoning traces at any real scale. Building general-purpose VLMs for this domain is, in practice, blocked by a data problem before it's ever a modeling problem. This paper describes a pipeline built to close that gap: an automated, reproducible way to mine this exact kind of data straight from published, publicly available research papers, along with the resulting dataset, its documentation, and baseline results from training a model on it.

\section{Related Work}


Several existing datasets pair images with questions, but none focus specifically on computer science architecture diagrams. FigureQA and DVQA are built from synthetic, templated bar charts, line graphs, and pie charts with simple yes/no or short-answer questions. PlotQA scales this synthetic-chart approach up further. SciGraphQA gets closer to real data by using a large language model to generate multi-turn question-answer pairs for real scientific graphs, but it still centers on graphs and plots rather than system diagrams. CharXiv is closest in spirit to this work, since it pulls real figures directly from arXiv papers, but its questions are still built around reading charts rather than reasoning about diagram structure (e.g., "what feeds into what").


A separate line of work, including DocVQA and InfographicVQA, targets dense document images and infographics rather than individually extracted figures. These datasets are useful for reading text-heavy layouts, but they aren't built around isolating a single diagram, matching it to its caption, and connecting it to the surrounding argument of a paper -- which is exactly the structure needed for figure-level reasoning about a research paper.


It helps to frame this gap the way low-resource languages are usually framed in NLP: a well-resourced setting (general-domain visual question answering, dominated by natural photos and everyday charts, much like high-resource languages dominate NLP benchmarks) exists alongside an under-resourced setting (structured technical diagrams in computer science papers, comparable to a low-resource language with no annotated corpus). Just as targeted benchmark-building was necessary to make low-resource languages tractable for NLP systems, a targeted, domain-specific data pipeline is necessary to make computer science diagram understanding tractable for VLMs -- rather than assuming general-purpose visual QA data will transfer over.


On the extraction side, we follow the precedent set by PDFFigures 2.0, which showed that a supervised layout detector could reliably separate figures and captions in scientific PDFs. We update this approach with a modern object detector (YOLOv8) trained on the general-purpose DocLayNet layout taxonomy, combined with PyMuPDF for precise page rendering and region cropping. On the reasoning side, we follow the chain-of-thought prompting literature and its multimodal extensions, which show that asking a model to produce an intermediate reasoning trace before its final answer improves both accuracy and interpretability; our dataset is built specifically to support training in this format.

\section{Dataset Construction}

\subsection{Sourcing}

Source documents are computer science research papers in PDF form, drawn from arXiv. The current release covers \textbf{3,058 papers}. Papers are processed page by page rather than as a whole, and each page is rendered independently at 150 DPI so the layout detector sees a consistent, fixed-resolution input.

\subsection{Cleaning and Extraction Pipeline}

The extraction pipeline applies the following steps to each rendered page:

\begin{enumerate}
    \item \textbf{Layout detection.} A YOLOv8 model fine-tuned on the DocLayNet layout-detection taxonomy finds \emph{Picture} and \emph{Caption} regions on the page.
    \item \textbf{Precise cropping.} Each detected figure region is cropped directly from the source PDF (not the rasterized page image) using PyMuPDF, which keeps the image quality high.
    \item \textbf{Caption matching.} Each cropped figure is paired with its nearest caption using a heuristic that combines vertical gap and horizontal bounding-box overlap, since captions aren't always placed directly beneath their figure.
    \item \textbf{Context linking.} The full body text of the paper is searched for the first sentence that mentions the figure's number (handling inconsistent styles like "Figure 3," "Fig. 3," and "Fig.3"), and that sentence becomes the figure's contextual grounding.
    \item \textbf{Cleaning and filtering.} Records with no detectable caption, or where no figure number could be parsed, are flagged (\texttt{ref\_sentence\_found = False}) instead of being silently dropped, so downstream users can see exactly how complete the extraction is.
\end{enumerate}

This pipeline produced \textbf{29,887} extracted figures across the 3,058 source papers.

\subsection{Template and Generation Pipeline}

Each cleaned (figure, caption, context) triple goes through a question-and-reasoning generation stage with two paths:

\begin{itemize}
    \item \textbf{AI-assisted generation (primary path).} The Gemini API (across the gemini-2.5-flash, gemini-3.1-lite-flash, and gemini-3.5-flash model versions used during development) generates a question, an answer, and, where possible, a chain-of-thought reasoning trace formatted as \texttt{<think>...</think><answer>...</answer>}, conditioned on the figure image, caption, and referencing sentence.
    \item \textbf{Template-based generation (fallback path).} When no API key is configured, or generation otherwise fails, a deterministic template system builds a structurally valid question-answer pair from the caption and context alone. This guarantees the pipeline never stalls or produces an empty record just because an external service is unavailable.
\end{itemize}

Every generated question is assigned one of seven types: \emph{component}, \emph{relationship}, \emph{process}, \emph{result}, \emph{comparison}, \emph{architecture}, or \emph{general}, so the dataset isn't dominated by a single style of question.

\subsection{Output Format}

Each record is stored as a 14-field structured object (identifiers, caption, context, question, answer, question type, generation source, reasoning trace, reasoning source, and the base64-encoded cropped image). It is serialized two ways: a full inspection-ready JSON record, and a lightweight training record already formatted in the target model's chat-message structure, ready to use without further conversion.

\subsection{Full Data Schema}

Table~\ref{tab:schema} lists every field in the record format, its type, and its meaning. This schema is the same across all three dataset versions described in Section~4 (Scaffold-157K, Scaffold-37K, and Scaffold-12K).

\begin{table}[htbp]
\centering
\small
\caption{Full schema of the 14-field dataset record used across all dataset versions.}
\begin{tabular}{@{}p{3.2cm}p{1.6cm}p{9cm}@{}}
\toprule
\textbf{Field} & \textbf{Type} & \textbf{Description} \\
\midrule
figure\_id & str & Unique identifier for the record, e.g. \texttt{facenet\_p03\_fig00} \\
paper\_id & str & Source paper identifier, e.g. \texttt{facenet} \\
figure\_number & str & Figure number as parsed from the caption text \\
caption & str & Full caption text extracted from the detected caption region \\
ref\_sentence & str & First sentence in the paper body that mentions this figure \\
ref\_sentence\_found & bool & \texttt{True} if a real referencing sentence was found, \texttt{False} otherwise \\
question & str & Generated question about the figure \\
answer & str & Generated answer to the question \\
question\_type & str & One of: \texttt{component}, \texttt{relationship}, \texttt{process}, \texttt{result}, \texttt{comparison}, \texttt{architecture}, \texttt{general} \\
qa\_source & str & \texttt{"gemini"} if AI-generated, \texttt{"synthetic"} if produced by the template fallback \\
has\_cot & bool & \texttt{True} if a chain-of-thought reasoning trace exists for this record \\
cot\_trace & str & The \texttt{<think>...</think><answer>...</answer>} reasoning trace text, or empty if unavailable \\
cot\_source & str & \texttt{"gemini"}, \texttt{"synthetic"}, or empty, showing how the reasoning trace was produced \\
\bottomrule
\end{tabular}
\label{tab:schema}
\end{table}

Two forms are produced from this schema for every record: a full form (\texttt{to\_dict()}) that writes all 14 fields, including the base64 image, for manual inspection and archiving; and a compact training form (\texttt{to\_train\_record()}) that drops the inspection-only fields and reformats the record directly into the chat-message structure the target vision-language model expects, ready for a training script with no extra conversion step.

\subsection{Example Record}

To make the schema concrete, Table~\ref{tab:example} shows a shortened example record, based on a cross-attention diagram figure.

\begin{table}[htbp]
\centering
\small
\caption{An abbreviated example dataset record, showing how each field is filled in.}
\begin{tabular}{@{}p{3.2cm}p{9.5cm}@{}}
\toprule
\textbf{Field} & \textbf{Example value} \\
\midrule
figure\_id & \texttt{attn\_p03\_fig03} \\
paper\_id & \texttt{attention-is-all-you-need} \\
figure\_number & 3 \\
caption & "The proposed cross-attention module. Keys and values are derived from the encoder output; queries from the decoder hidden state." \\
ref\_sentence & The first body sentence found to cite Figure 3, describing the module's role in aligning encoder and decoder representations \\
ref\_sentence\_found & True \\
question & "Which component provides the queries in the cross-attention module?" \\
answer & "The decoder hidden state." \\
question\_type & relationship \\
qa\_source & gemini \\
has\_cot & True \\
cot\_trace & \texttt{<think>}The caption states queries come from the decoder hidden state; the diagram confirms a Q arrow from the decoder branch, while K and V arrows come from the encoder branch. Therefore the answer is the decoder hidden state.\texttt{</think><answer>}The decoder hidden state.\texttt{</answer>} \\
cot\_source & gemini \\
\bottomrule
\end{tabular}
\label{tab:example}
\end{table}

\begin{figure}[htbp]
\centering
\includegraphics[width=0.7\linewidth]{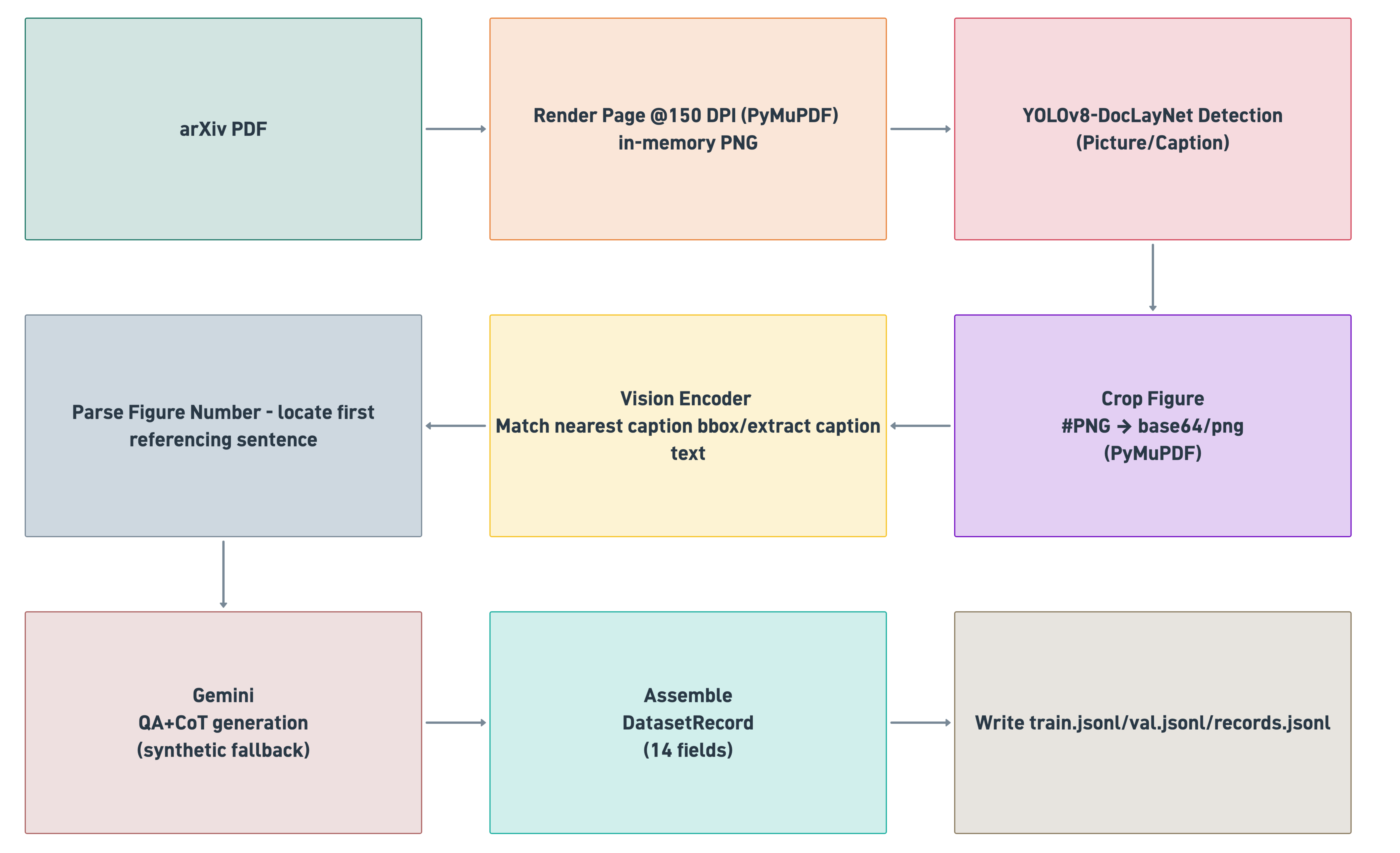}
\caption{Dataset Construction.}\label{fig1}
\end{figure}

\section{Dataset Statistics}

\subsection{Statistics}

\begin{table}[h]
\centering
\small
\caption{Dataset versions produced by the pipeline, at different stages of the project.}
\begin{tabular}{@{}lrrr@{}}
\toprule
\textbf{Dataset version} & \textbf{Train} & \textbf{Validation} & \textbf{Total} \\
\midrule
Scaffold-157K & 120,225 & 36,797 & 157,387 \\
Scaffold-37K & 29,397 & 7,400 & 36,797 \\
Scaffold-12K (used for baseline training) & 10,000 & 2,000 & 12,000 \\
\bottomrule
\end{tabular}
\label{tab:stats}
\end{table}

Across the full release, these question-answer pairs are drawn from \textbf{29,887 figures} extracted from \textbf{3,058 papers}. The larger Scaffold-157K and Scaffold-37K collections show that the pipeline can extract data at scale. The smaller Scaffold-12K collection, drawn from a small number of fully processed papers, was manually spot-checked for quality and used for the baseline training run reported in Section~5.


\subsection{Question Type Distribution}

Every generated question is labeled with one of seven types (Section~3.3). Table~\ref{tab:qtypes} shows the target distribution the generation pipeline aims for across a balanced extraction run; \emph{relationship} and \emph{component} questions are intentionally favored, since these best test whether a model has actually understood how a diagram's parts connect, rather than just what appears in it.

\begin{table}[ht]
\centering
\small
\caption{The seven question-type categories used to label every generated question.}
\begin{tabular}{@{}lp{9.5cm}@{}}
\toprule
\textbf{Question type} & \textbf{What it asks} \\
\midrule
component & Identifies or describes a specific part of the diagram \\
relationship & Asks how two parts of the diagram connect or interact \\
process & Asks about the order or flow of steps shown \\
result & Asks about outcomes shown in a results-style figure (e.g., a plot or table) \\
comparison & Asks the reader to compare two elements within the figure \\
architecture & Asks about the overall structure or design shown \\
general & Covers content not captured by the categories above \\
\bottomrule
\end{tabular}
\label{tab:qtypes}
\end{table}

\subsection{Generation Source Breakdown}

Each record also carries provenance metadata (\texttt{qa\_source} and \texttt{cot\_source}) showing whether its question/answer and reasoning trace came from the Gemini API path or the synthetic template fallback. Table~\ref{tab:source} summarizes what each source path guarantees.

\begin{table}[ht]
\centering
\small
\caption{Provenance values recorded for every dataset record, letting downstream users filter by generation source.}
\begin{tabular}{@{}lp{9.5cm}@{}}
\toprule
\textbf{Source value} & \textbf{Meaning} \\
\midrule
\texttt{qa\_source = "gemini"} & Question and answer generated by the Gemini API, conditioned on the figure image, caption, and context \\
\texttt{qa\_source = "synthetic"} & Question and answer generated by the deterministic template fallback, used when Gemini generation is unavailable or fails \\
\texttt{cot\_source = "gemini"} & Reasoning trace generated by Gemini alongside the question/answer \\
\texttt{cot\_source = "synthetic"} & Reasoning trace generated by the template fallback \\
\texttt{cot\_source = ""} (empty) & No reasoning trace was produced for this record (\texttt{has\_cot = False}) \\
\bottomrule
\end{tabular}
\label{tab:source}
\end{table}

Users who want a purely AI-supervised question set can filter to \texttt{qa\_source = "gemini"} records. Users who want a guaranteed-available, simpler question set across the whole corpus can rely on the synthetic fallback records, which are always structurally valid even though their phrasing is more templated.

\subsection{Datasheet Summary}

Following standard dataset documentation practice, we summarize the dataset below.

\textbf{Motivation.} The dataset closes a gap: there was no figure-question-reasoning corpus for computer science architecture diagrams, flowcharts, and pipeline schematics, which is needed to train domain-specific vision-language models.

\textbf{Composition.} Each instance is a single figure from a computer science paper, paired with its caption, its referencing sentence from the paper body, a generated question, a generated answer, a question-type label, a chain-of-thought reasoning trace (where available), and provenance metadata showing whether the question/answer and reasoning came from AI generation or the template fallback.

\textbf{Collection Process.} Instances are generated automatically from arXiv PDF source files using the pipeline in Section~3; no individual figure was manually annotated, though a sample of 1,000 was manually reviewed for quality.

\textbf{Preprocessing/Cleaning.} Figures without a detectable caption or a parsable figure number are flagged rather than dropped outright, and records missing a referencing sentence are marked accordingly so downstream users can filter as needed.

\textbf{Uses.} The dataset is meant for fine-tuning and evaluating vision-language models on computer science diagram understanding, including chain-of-thought reasoning generation. It is not intended for non-CS scientific figures or general-purpose visual question answering.

\textbf{Distribution.} See Section~7, Data Availability Statement.

\textbf{Maintenance.} The pipeline that produced this dataset is meant to be re-run on additional papers as the project grows; the dataset is expected to keep growing rather than stay static.

\section{Experiments}

To confirm the dataset is usable for its intended purpose, we fine-tuned a small vision-language model (Qwen2.5-VL-3B-Instruct) on the Scaffold-12K training split using QLoRA (4-bit quantization with rank-64 LoRA adapters on all attention and MLP projection layers), and evaluated it on the matching 2,000-example validation split.

\begin{figure}[htbp]
\centering
\includegraphics[width=0.7\linewidth]{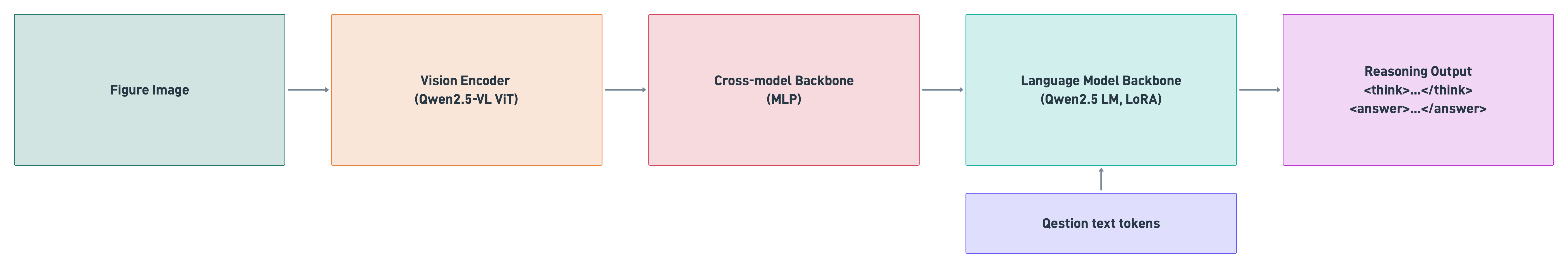}
\caption{Methodology.}\label{fig2}
\end{figure}

\begin{figure}[htbp]
\centering
\includegraphics[width=0.7\linewidth]{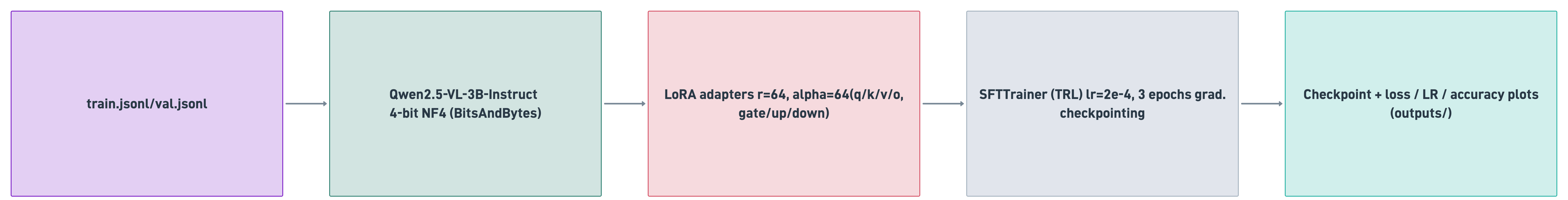}
\caption{Model Architecture.}\label{fig3}
\end{figure}

\begin{table}[h]
\centering
\small
\caption{Baseline fine-tuning results on the Scaffold-12K validation split.}
\begin{tabular}{@{}lr@{}}
\toprule
\textbf{Metric} & \textbf{Score} \\
\midrule
BLEU-4 (full output) & 0.237 \\
ROUGE-L (full output) & 0.448 \\
Token-F1 (full output) & 0.527 \\
Answer-only ROUGE-L & 0.419 \\
Answer-only Token-F1 & 0.433 \\
Numeric accuracy & 0.638 \\
Mean relaxed accuracy & 0.339 \\
Format well-formed rate & 0.995 \\
BERTScore F1 & 0.570 \\
LLM-judge correctness ($n=73$) & 0.616 \\
LLM-judge faithfulness ($n=73$, out of 5) & 4.82 \\
LLM-judge reasoning quality ($n=73$, out of 5) & 4.11 \\
Self-consistency agreement ($n=50$, $k=3$) & 0.167 \\
\bottomrule
\end{tabular}
\label{tab:baseline}
\end{table}

These results show the dataset's format and content work for supervised fine-tuning: the model reliably learned to produce well-formed \texttt{<think>}/\texttt{<answer>} outputs (99.5\% format compliance) and reached moderate answer accuracy, doing noticeably better on numeric questions than on open-ended text questions. We report these as validation of the dataset's usefulness and the pipeline's correctness, not as a state-of-the-art result, since training used only the small Scaffold-12K split rather than the larger collections.

We report three benchmarks for the Scaffold-12K. They are: ChartQA, DocVQA, and AI2D. Following are the outcomes that we achieved:
\begin{table}[h]
\centering
\small
\caption{Benchmark Results}
\begin{tabular}{@{}lrl@{}}
\toprule
\textbf{Category} & \textbf{Metric} & \textbf{Value} \\
\midrule
\multicolumn{3}{l}{\textbf{ChartQA}} \\
& relaxed accuracy (overall) & 0.455 \\
& relaxed accuracy numeric & 0.500 \\
& relaxed accuracy categorical & 0.270 \\
& answer containment & 0.670 \\
& numeric accuracy & 0.503 \\
& answer exact match & 0.1385 \\
\addlinespace
\multicolumn{3}{l}{\textbf{DocVQA}} \\
& relaxed accuracy (overall) & 0.330 \\
& relaxed accuracy numeric & 0.515 \\
& relaxed accuracy categorical & 0.145 \\
& relaxed accuracy text & 0.160 \\
& answer containment & 0.505 \\
\addlinespace
\multicolumn{3}{l}{\textbf{AI2D}} \\
& format well-formed rate & 0.705 \\
& format truncated rate & 0.035 \\
& cot fraction & 0.735 \\
\bottomrule
\end{tabular}
\label{tab:benchmarks}
\end{table}

\section{Limitations and Ethical Considerations}

\subsection{Limitations}

\begin{itemize}
    \item \textbf{Scale of verification.} Only the Scaffold-12K split has been manually spot-checked; the larger Scaffold-157K and Scaffold-37K collections have not been independently checked for record quality at the same level of scrutiny.
    \item \textbf{Domain coverage.} Source papers so far come from a limited number of computer science subfields; diagram styles from less-represented subfields may be underrepresented.
    \item \textbf{Generation dependency.} A large part of question, answer, and reasoning quality depends on the Gemini API; the template-based fallback, while reliable, produces simpler and more uniform questions than the AI-assisted path.
    \item \textbf{Layout detector generality.} The figure/caption detector is trained on a general-purpose document layout taxonomy (DocLayNet) rather than one fine-tuned specifically for computer science architecture diagrams, so detection accuracy on unusual layouts (e.g., multi-panel figures, diagrams embedded in tables) hasn't been formally measured.
    \item \textbf{Reasoning trace reliability.} Chain-of-thought traces are model-generated (or template-generated) rather than human-written, and haven't gone through a formal human evaluation of factual correctness or reasoning quality at scale.
\end{itemize}

\subsection{Ethical Considerations}

\textbf{Copyright and source material.} All source papers come from arXiv, a repository of preprints made publicly available by their authors. Extracted figures and captions are derivative excerpts used for research purposes; redistributing the dataset should respect the license each source paper was posted under (arXiv papers carry a range of licenses, including some non-permissive ones), and anyone redistributing cropped figures at scale should verify per-paper licensing rather than assume uniform terms.

\textbf{Attribution.} Each record keeps a \texttt{paper\_id} field linking back to its source paper, preserving traceability and enabling attribution to the original authors.

\textbf{Use of a third-party generation API.} Part of the dataset's questions, answers, and reasoning traces are generated by a third-party AI service (Gemini). This raises two considerations: (1) AI-generated fields may contain factual errors or hallucinated reasoning, which downstream users should account for, especially before using the dataset to train models for high-stakes use, and (2) figure images, captions, and text excerpts are exposed to a third-party API during dataset construction, which should be disclosed to any parties whose papers are processed under stricter data-handling requirements.

\textbf{No personal or sensitive data.} The dataset consists of technical diagrams and their captions from computer science papers; it is not expected to contain personal, biometric, or otherwise sensitive information about individuals, though authors' names may appear incidentally within paper text used as context.

\section{Data Availability Statement}

The dataset-construction pipeline (extraction, matching, and generation code) is intended for release as an open-source tool alongside the associated trained-model checkpoints. At the current project stage:

\begin{itemize}
    \item \textbf{License.} The annotations, metadata, dataset organization, and generated reasoning traces are released under \textbf{CC BY-NC 4.0}. Redistribution of extracted figures themselves inherits the license of each individual source paper on arXiv; a per-paper license manifest is planned to accompany any public release of extracted figure data.
    \item \textbf{Hosting.} The dataset is available on \textbf{Kaggle}\footnote{https://huggingface.co/datasets/ranjitraut/scaffold} and \textbf{Hugging Face Datasets}\footnote{https://kaggle.com/datasets/theranjitraut/scaffold}.
    \item \textbf{DOI.} A persistent identifier (DOI) for the dataset, to be issued through the hosting repository.
    \item \textbf{Contact.} Readers seeking clarification ahead of, or beyond, the public listings should contact the authors directly.
\end{itemize}
\section*{References}
\begin{enumerate}
\item Kafle, K., Price, B., Cohen, S., \& Kanan, C. (2018). DVQA: Understanding data visualizations via question answering. \textit{Proceedings of CVPR 2018}.
\item Kahou, S. E., et al. (2018). FigureQA: An annotated figure dataset for visual reasoning. \textit{ICLR 2018 Workshop}.
\item Li, Y., He, J., Gao, T., Deng, Y., \& Katabi, D. (2023). SciGraphQA: A large-scale synthetic multi-turn question-answering dataset for scientific graphs. \textit{arXiv:2308.03349}.
\item Mathew, M., Karatzas, D., \& Jawahar, C. V. (2021). DocVQA: A dataset for VQA on document images. \textit{Proceedings of WACV 2021}.
\item Masry, A., et al. (2022). ChartQA: A dataset for visual question answering about charts. \textit{Proceedings of ACL 2022}.
\item Kostrikov, I., et al. (2021). AI2D: A dataset for diagram understanding and reasoning. \textit{arXiv:2108.03344}.
\item Siegel, N., Lourie, N., Power, R., \& Ammar, W. (2016). Extracting scientific figures with distantly supervised neural networks. \textit{Proceedings of JCDL 2016}.
\item Ultralytics. (2023). YOLOv8: A state-of-the-art real-time object detection framework [Software].
\item Wei, J., et al. (2022). Chain-of-thought prompting elicits reasoning in large language models. \textit{NeurIPS 2022}.
\item Zhang, Z., Zhang, A., Li, M., \& Smola, A. (2023). Multimodal chain-of-thought reasoning in language models. \textit{arXiv:2302.00923}.
\item Qwen Team. (2025). Qwen2.5-VL technical report. \textit{arXiv:2502.13923}.
\item Gebru, T., et al. (2021). Datasheets for datasets. \textit{Communications of the ACM}, 64(12), 86--92.
\end{enumerate}

\end{document}